\documentclass[lettersize,journal]{IEEEtran}
\usepackage{tikz,xcolor,hyperref}
\definecolor{lime}{HTML}{A6CE39}
\DeclareRobustCommand{\orcidicon}{%
    \begin{tikzpicture}
    \draw[lime, fill=lime] (0,0) 
    circle [radius=0.16] 
    node[white] {{\fontfamily{qag}\selectfont \tiny ID}};    \draw[white, fill=white] (-0.0625,0.095) 
    circle [radius=0.007];    \end{tikzpicture}
    \hspace{-2mm}}
\foreach \x in {A, ..., Z}{%
    \expandafter\xdef\csname orcid\x\endcsname{\noexpand\href{https://orcid.org/\csname orcidauthor\x\endcsname}{\noexpand\orcidicon}}
    }

\usepackage{amsmath,amsfonts}
\usepackage{algorithm}
\usepackage{algpseudocode}
\usepackage{array}
\usepackage{textcomp}
\usepackage{stfloats}
\usepackage{url}
\usepackage{verbatim}
\usepackage{graphicx}
\usepackage{cite}
\usepackage{amsmath}
\usepackage{amssymb}
\usepackage{mathtools}
\usepackage{amsthm}
\usepackage{hyperref}
\usepackage{url}
\usepackage[utf8]{inputenc} 
\usepackage[T1]{fontenc}    
\usepackage{hyperref}       
\usepackage{url}            
\usepackage{booktabs}       
\usepackage{amsfonts}       
\usepackage{nicefrac}       
\usepackage{microtype}      
\usepackage[table]{xcolor}        
\usepackage{graphicx}
\usepackage{wrapfig}
\usepackage{multirow}
\usepackage{soul}
\usepackage{makecell}

\usepackage{subcaption}
\usepackage{comment}
\usepackage{threeparttable}
\usepackage{float}
\usepackage{bm}
\usepackage[english]{babel}
\def\BibTeX{{\rm B\kern-.05em{\sc i\kern-.025em b}\kern-.08em
    T\kern-.1667em\lower.7ex\hbox{E}\kern-.125emX}}
\usepackage{balance}

\definecolor{lightblue}{rgb}{0.21, 0.49, 0.74}

\begin{document}
\title{UniDiff: A Unified Diffusion Framework for Multimodal Time Series Forecasting}

\author{Da~Zhang\orcidA{},~\IEEEmembership{Student Member,~IEEE,} 
Bingyu~Li\orcidB{}, 
Zhiyuan~Zhao\orcidC{},
Junyu~Gao\orcidD{},~\IEEEmembership{Member,~IEEE,} \\ Feiping~Nie\orcidE{},~\IEEEmembership{Senior Member,~IEEE,}~and~Xuelong~Li\orcidF{},~\IEEEmembership{Fellow,~IEEE}

\thanks{Da Zhang, Junyu Gao, and Feiping Nie are with the School of Artificial Intelligence, OPtics and ElectroNics (iOPEN), Northwestern Polytechnical University, Xi'an 710072, China and also with the Institute of Artificial Intelligence (TeleAI), China Telecom, China. (E-mail: dazhang@mail.nwpu.edu.cn; gjy3035@gmail.com; feipingnie@gmail.com).}

\thanks{Bingyu Li, Zhiyuan Zhao, and Xuelong Li are with the Institute of Artificial Intelligence (TeleAI), China Telecom, China. (E-mail: libingyu0205@mail.ustc.edu.cn; tuzixini@gmail.com; xuelong\_li@ieee.org).}

\thanks{Junyu Gao and Xuelong Li are corresponding authors.}
}


\maketitle

\begin{abstract}
As multimodal data proliferates across diverse real-world applications, leveraging heterogeneous information such as texts and timestamps for accurate time series forecasting (TSF) has become a critical challenge. 
While diffusion models demonstrate exceptional performance in generation tasks, their application to TSF remains largely confined to modeling single-modality numerical sequences, overlooking the abundant cross-modal signals inherent in complex heterogeneous data. 
To address this gap, we propose UniDiff, a unified diffusion framework for multimodal time series forecasting. To process the numerical sequence, our framework first tokenizes the time series into patches, preserving local temporal dynamics by mapping each patch to an embedding space via a lightweight MLP. At its core lies a unified and parallel fusion module, where a single cross-attention mechanism adaptively weighs and integrates structural information from timestamps and semantic context from texts in one step, enabling a flexible and efficient interplay between modalities. Furthermore, we introduce a novel classifier-free guidance mechanism designed for multi-source conditioning, allowing for decoupled control over the guidance strength of textual and temporal information during inference, which significantly enhances model robustness. Extensive experiments on real-world benchmark datasets across eight domains demonstrate that the proposed UniDiff model achieves state-of-the-art performance.

\end{abstract}

\begin{IEEEkeywords}
Multimodal Fusion; Time Series; Forecasting; Diffusion Model
\end{IEEEkeywords}

\section{Introduction}
\label{sec:intro}

\IEEEPARstart{T}{ime} series forecasting (TSF), the task of predicting future values based on historical observations \cite{cheng2025comprehensive}, is a cornerstone of data-driven decision-making across a multitude of domains in modern data-driven ecosystems \cite{liu2024unitime, huang2025exploiting}. 
From forecasting e-commerce sales \cite{kalifa2022leveraging} and predicting web traffic flow \cite{zhang2025mamnet} to analyzing financial markets \cite{arsenault2025survey} influenced by online news and reports, accurate TSF is critical for optimizing resources, mitigating risks, and strategic planning. 
With the proliferation of deep learning, TSF research has shifted from classical statistical models towards sophisticated neural networks \cite{song2024deep}. Architectures such as Recurrent Neural Networks (RNNs) \cite{lin2023segrnn} and Transformers \cite{Yuqietal-2023-PatchTST} have demonstrated superior capacity in modeling complex temporal dependencies within numerical sequences.

\begin{figure}[t]
	\centering
	\includegraphics[width=1\linewidth]{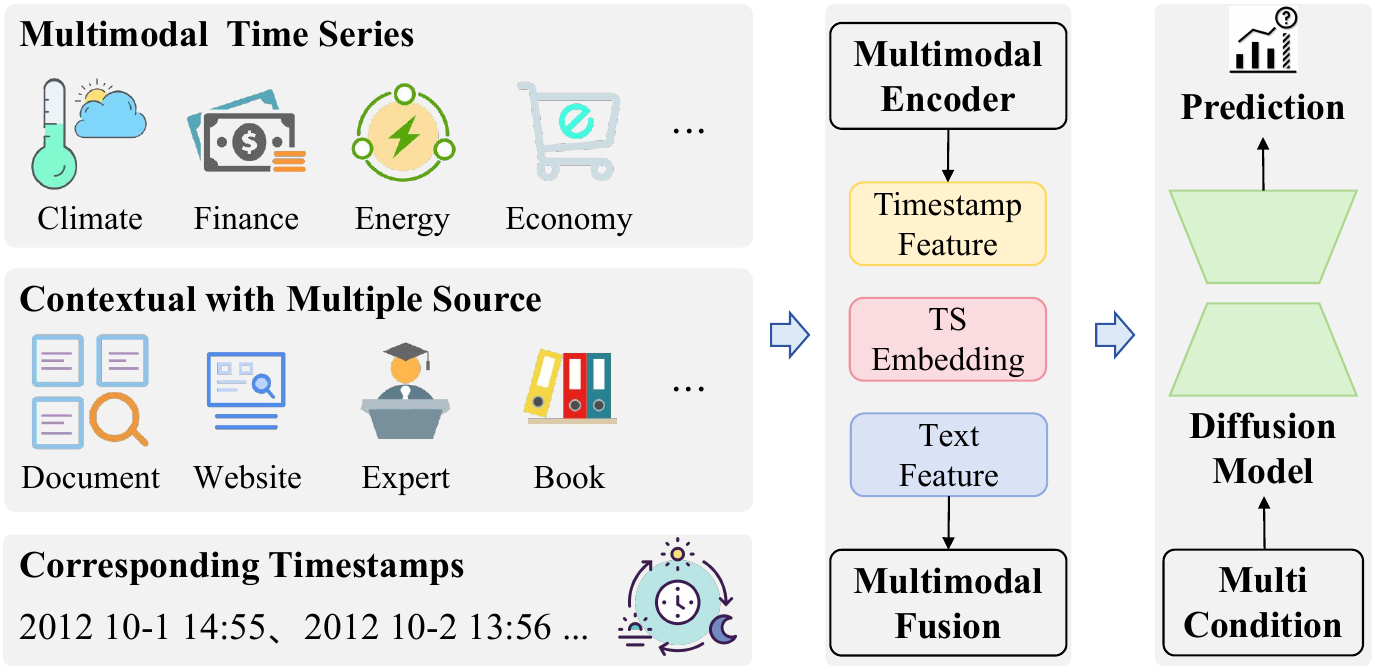}
	\caption{The landscape of Multimodal TSF. Our proposed UniDiff framework processes core numerical series alongside rich contextual information from texts and structural cues from timestamps to generate robust predictions.}
	\label{fig1}

\end{figure}

Whereas, relying solely on historical numerical data presents a fundamental limitation: such data often lacks the rich contextual information necessary to explain and predict shifts driven by external events \cite{wang2024news, xu2024beyond}. 
Data-intensive real-world applications are prime examples where auxiliary, non-numeric data is abundant and influential \cite{koh2024visualwebarena}. 
For instance, a product's sales volume may be significantly impacted by social media trends and news articles, while stock prices often react to financial reports and expert analyses. 
This has spurred a growing interest in multimodal TSF, which seeks to integrate heterogeneous data sources, particularly text and timestamps, to build more comprehensive and accurate forecasting models \cite{liu2024time, zhang2024multivariate, jiang2025multi}. 
As illustrated in Figure \ref{fig1}, such tasks require the integration of core numerical series from different domains (e.g., finance, energy), contextual text from diverse sources (e.g., news, reports), and corresponding timestamps that provide structural information, all to achieve more accurate predictions.

\begin{figure*}[t]
	\centering
	\includegraphics[width=1\linewidth]{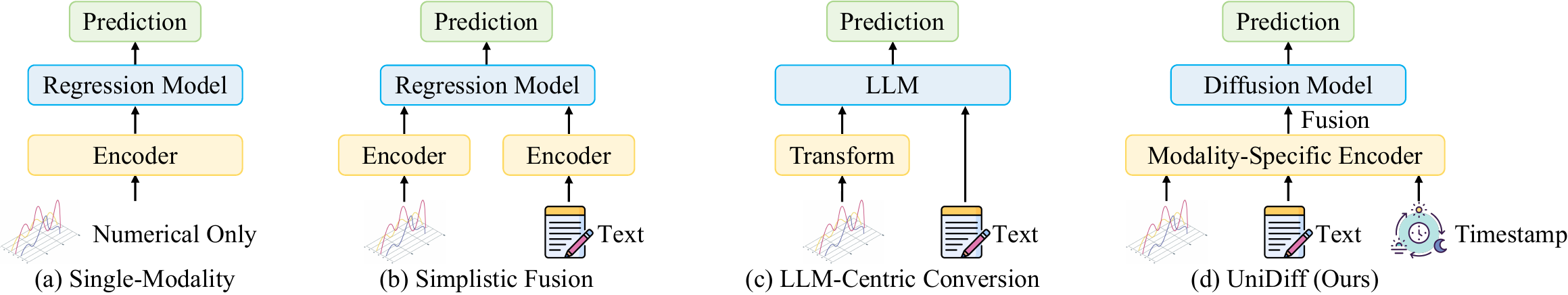}
	\caption{Different architectural for TSF. (a) Single-modality models rely solely on historical numerical data, ignoring rich contextual information. (b) Simplistic fusion models use rudimentary techniques like concatenation, failing to capture complex inter-modal dynamics. (c) LLM-centric conversion models transform all data into text, risking the loss of precision and inherent characteristics of numerical data. (d) our proposed UniDiff framework employs a unified and parallel fusion mechanism to adaptively integrate numerical, temporal, and textual information, overcoming the limitations of prior architectures.}
	\label{fig2}

\end{figure*}

The integration of these diverse modalities, however, introduces significant challenges. Early approaches often struggled with effective fusion strategies \cite{chattopadhyay2024context, kim2024multi}. 
More recent works leveraging Large Language Models (LLMs) have shown promise in processing textual information \cite{jin2023time, wang2025chattime, su2025fusing}. Yet, some methods resort to converting numerical time series into textual formats for the LLM to process, a strategy that risks losing the inherent continuous characteristics and precision of the original data \cite{kong2025position}. 
Furthermore, many of these models treat the time series as a deterministic input, failing to account for the inherent randomness and volatility that characterize real-world data \cite{jia2024gpt4mts}. 
This discrepancy often leads to a performance bottleneck and calls for solutions that can explicitly model this uncertainty, such as probabilistic frameworks \cite{shen2023non, yang2024survey}.

In response, diffusion models have recently emerged as a powerful probabilistic paradigm for TSF \cite{tashiro2021csdi, li2022generative, liu2025stochastic}. 
By framing forecasting as an iterative denoising process conditioned on historical data, these models excel at capturing the stochastic nature of time series \cite{shen2024multi}.
Despite their potential, the application of diffusion models to multimodal TSF is still in its nascent stages, with existing approaches often adopting simplistic conditioning mechanisms \cite{lin2024diffusion}. 
For instance, many current methods either directly concatenate multimodal features \cite{wang2024multi} or employ basic fusion techniques that lack the sophistication to capture the complex, dynamic interplay between different data streams \cite{su2025multimodal}. 
Such rudimentary fusion strategies fail to allow the model to dynamically arbitrate the relative importance of each modality during the denoising process \cite{narasimhan2024time}. .
Moreover, vital control mechanisms like classifier-free guidance (CFG), which are crucial for improving sample quality, have thus far been designed to operate on a single conditional modality, limiting their flexibility and effectiveness in a true multimodal setting \cite{ijcai2025p580}. Figure \ref{fig2} provides a comparative illustration of these existing architectures against our proposed approach.

In this paper, we address these fundamental limitations by proposing UniDiff, a unified diffusion framework for multimodal TSF. UniDiff is architected to perform a more flexible and synergistic fusion of numerical, temporal, and textual information. To effectively capture local patterns, our framework first tokenizes the time series into patches and utilizes a lightweight MLP to project them into an embedding space, preserving local temporal dynamics. The cornerstone of UniDiff is a unified and parallel fusion module. Unlike simplistic concatenation or rigid, sequential methods, this module employs a single cross-attention mechanism to concurrently process embeddings from the time series, timestamps, and text. This parallel architecture empowers the model to adaptively weigh the influence of each modality at every denoising step, enabling a more direct and efficient interplay between them. Furthermore, to enhance robustness and controllability, we introduce a novel CFG mechanism designed for multi-source conditioning, which allows for decoupled guidance from both timestamp and text modalities. This approach enables fine-grained, independent control over the influence of each information source during inference, mitigating the negative impact of potentially noisy or irrelevant conditional inputs.
The main contributions of our work are summarized as follows:

\begin{itemize}
\item We propose UniDiff, a novel unified diffusion framework for multimodal TSF that performs parallel and adaptive fusion of numerical, temporal, and textual data.
\item A unified cross-attention fusion module is designed to facilitate dynamic, one-step integration of multiple modalities.
\item we introduce a CFG mechanism designed for multi-source conditioning, which allows for decoupled guidance from both timestamp and text modalities.
\item Extensive experiments on nine datasets demonstrate that UniDiff achieves state-of-the-art performance, validating its effectiveness.
\end{itemize}

\begin{figure*}[t]
	\centering
	\includegraphics[width=1\linewidth]{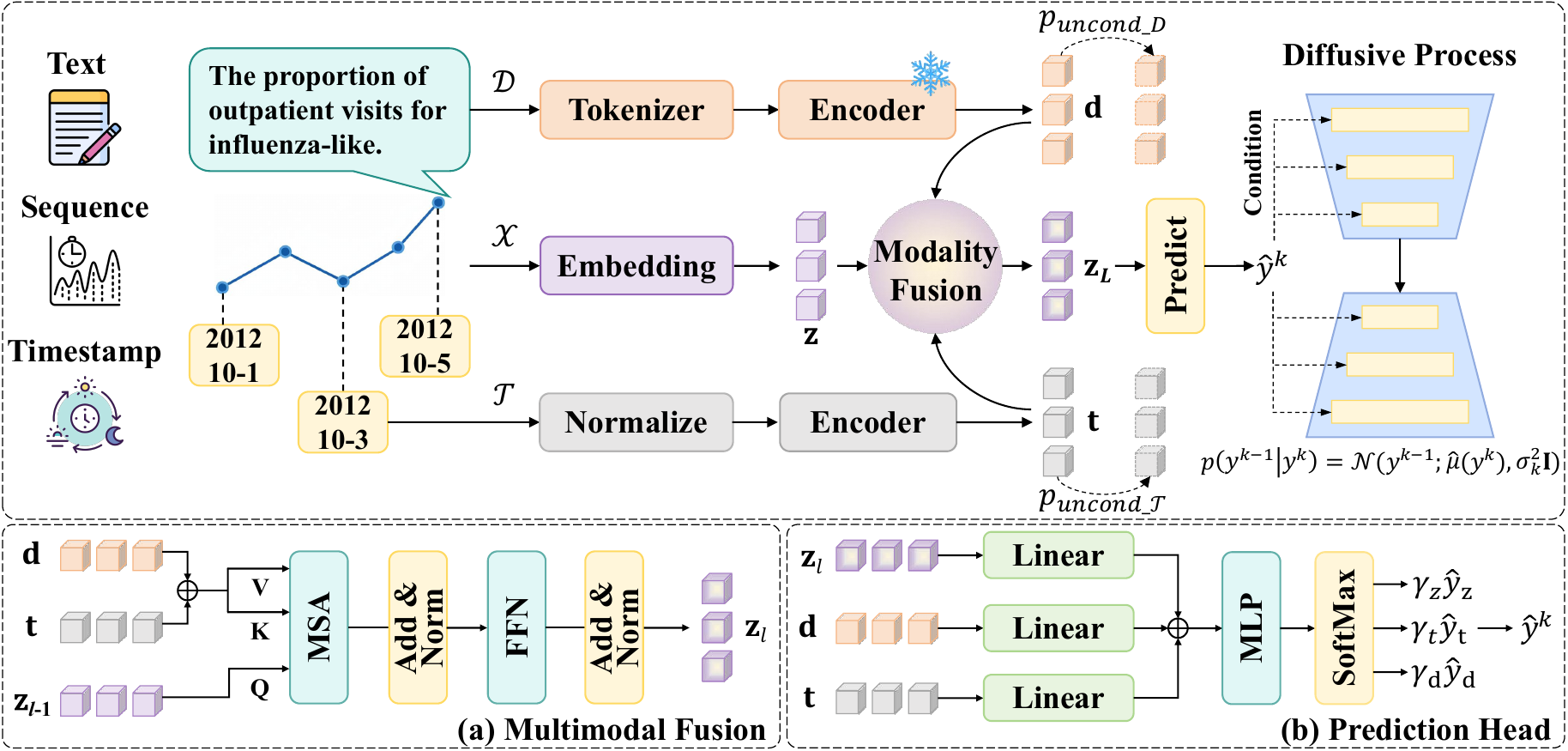}
	\caption{The overall architecture of the proposed UniDiff framework. The model processes three input modalities: Text ($\mathcal{D}$), the time series Sequence ($\mathcal{X}$), and Timestamps ($\mathcal{T}$). These inputs are first encoded into feature representations ($d$, $z$, and $t$). The core of the model consists of: \textbf{(a)} a \textbf{Multimodal Fusion} module that uses a unified cross-attention mechanism to integrate textual and temporal context into the sequence representation, and \textbf{(b)} a \textbf{Prediction Head} that generates the denoised estimate for the current step. This estimate conditions the iterative \textbf{Diffusive Process} to produce the final forecast.}
	\label{fig3}

\end{figure*}

\section{Related Work}

\subsection{Multimodal Time Series Forecasting}

The integration of auxiliary data to enhance TSF has become a significant research direction, with text and timestamps being particularly prominent modalities \cite{liu2024time, su2025multimodal, zhou2025motime}. 
Recent works have increasingly leveraged Large Language Models (LLMs) to process textual information, either by using them as powerful text encoders \cite{jin2023time} or by converting numerical series into text-like formats to fit the model's architecture \cite{liu2025timecma}. 
Fusion with time series features is then typically handled by mechanisms like cross-attention to capture cross-modal dependencies \cite{hu2025contextalignment, jeong2024precyse}. 
However, many of these approaches are deterministic, which limits their ability to model the inherent uncertainty and stochasticity of real-world time series \cite{oh2024stable}.
Furthermore, their fusion strategies are often simplistic, such as direct feature concatenation \cite{liu2024mtsa, kong2025position}, or employ rigid, sequential pipelines that restrict the dynamic interplay between different modalities \cite{wu2025aurora, park2025unicast}. 
Figure \ref{fig2} visually summarizes these limitations, contrasting the existing paradigms with the parallel fusion architecture we introduce. 
Our work addresses this limitation by proposing a parallel fusion mechanism that allows for a more flexible and direct interplay between all conditioning modalities.

\subsection{Diffusion Models for Time Series}

Diffusion models have recently emerged as a powerful probabilistic paradigm for TSF, adept at capturing data uncertainty through an iterative denoising process conditioned on historical observations \cite{liu2024retrieval}. 
While many diffusion-based methods focus on conditioning using only historical time series data, the exploration of multimodal conditioning is still in its nascent stages \cite{gao2025auto, wu2025dual}. 
Current approaches that do incorporate multiple modalities often rely on rudimentary conditioning mechanisms \cite{tashiro2021csdi, crabbe2024time}. 
For instance, they might employ simple feature concatenation or adopt sequential fusion architectures within the denoising network, which imposes a restrictive, predefined hierarchy on cross-modal interactions \cite{ge2025t2s}. 
Moreover, vital control mechanisms like classifier-free guidance (CFG), which are crucial for enhancing generation quality, have historically been designed to operate on only a single conditional modality, typically text \cite{ma2024utsd, su2025multimodal}. 
This limits their flexibility in a true multi-source conditioning setting. 
Our work directly addresses these gaps by introducing a parallel fusion architecture and a novel, decoupled guidance mechanism designed for multiple modalities.


\section{Methodology}

The overall architecture of UniDiff is illustrated in Figure \ref{fig3}.
Given a historical time series $\mathcal{X} = \{x_1, \ldots, x_{L_{in}}\}$ of length $L_{in}$, along with its corresponding timestamps $\mathcal{T}$ and associated textual descriptions $\mathcal{D}$, our objective is to predict the future time series $\mathcal{Y} = \{x_{L_{in}+1}, \ldots, x_{L_{in}+L_{out}}\}$ of length $L_{out}$ for the future timestamps $\mathcal{T}'$. UniDiff models this task as a conditional generation problem, progressively synthesizing the future time series $\mathcal{Y}$ from a standard Gaussian noise vector $\boldsymbol{\epsilon}$ through a $K$-step iterative denoising process. This denoising process is represented by the sequence $\{\mathcal{Y}^K, \mathcal{Y}^{K-1}, \ldots, \mathcal{Y}^0\}$. In this sequence, $\mathcal{Y}^K$ is initialized from $\boldsymbol{\epsilon}$, and the final output $\mathcal{Y}^0$ serves as the prediction result $\hat{\mathcal{Y}}$.
The core of this framework is a denoising network, denoted as $f_\theta$, which is conditioned on a rich set of multimodal inputs. At each step $k$, the network takes the noisy future sequence $\mathcal{Y}^k$, the historical sequence $\mathcal{X}$, the complete set of timestamps $\mathcal{T}_{full} = [\mathcal{T}, \mathcal{T}']$, and the text $\mathcal{D}$ to predict a less noisy version, $\mathcal{Y}^{k-1}$. This process is distinguished by its unified parallel fusion mechanism, which adaptively integrates temporal and semantic information, and is further enhanced by a novel decoupled guidance strategy. After iteratively performing this denoising process for $K$ steps, we obtain the final prediction $\hat{\mathcal{Y}}$. In the following sections, we will first review the fundamentals of conditional diffusion models for time series forecasting and then elaborate on the specific details of the UniDiff framework and its core components.

\subsection{Preliminaries: Conditional Diffusion for TSF}

Standard diffusion models are a class of generative frameworks composed of two fundamental processes: a forward noising process and a reverse denoising process. In the forward process, the diffusion model iteratively injects Gaussian noise into a target time series $\mathcal{Y}$ over $K$ discrete timesteps. At an arbitrary diffusion step $k$, the noisy state $\mathcal{Y}^k$ is sampled from the target $\mathcal{Y}$ according to the following conditional probability distribution:
\begin{equation}
	p(\mathcal{Y}^k | \mathcal{Y}) = \mathcal{N}(\mathcal{Y}^k; \mu = \sqrt{\overline{\alpha}_k}\mathcal{Y},\ \Sigma = (1-\overline{\alpha}_k)I)
	\label{eq1}
\end{equation}
where $p(\cdot|\cdot)$ denotes the conditional probability density function, and $\mathcal{N}(x; \mu, \Sigma)$ represents a multivariate Gaussian distribution with mean $\mu$ and covariance matrix $\Sigma$. $I$ is an identity matrix with the same dimensions as $\mathcal{Y}$. Here, $\overline{\alpha}_k = \prod_{i=1}^k \alpha_i$ is the cumulative signal retention factor up to step $k$, where each $\alpha_i \in (0,1)$ represents the proportion of signal retained at step $i$.

Conversely, the reverse process aims to progressively reconstruct the original target $\mathcal{Y}$ from pure Gaussian noise $\mathcal{Y}^K$ by learning a parameterized denoising distribution:
\begin{equation}
	p(\mathcal{Y}^{k-1}|\mathcal{Y}^{k}) = \mathcal{N}(\mathcal{Y}^{k-1};\ \hat{\mu}(\mathcal{Y}^{k}),\ \sigma_{k}^{2}I)
\end{equation}
In this formula, $\sigma_{k}^{2}$ is a predefined variance corresponding to step $k$, and the mean $\hat{\mu}(\mathcal{Y}^{k})$ is calculated as:
\begin{equation}
	\hat{\mu}(\mathcal{Y}^{k}) = \frac{\sqrt{\overline{\alpha}_{k}}(1-\overline{\alpha}_{k-1})}{\sqrt{\overline{\alpha}_{k-1}}(1-\overline{\alpha}_{k})} \mathcal{Y}^{k} + \frac{\overline{\alpha}_{k-1}-\overline{\alpha}_{k}}{\sqrt{\overline{\alpha}_{k-1}}(1-\overline{\alpha}_{k})} \hat{\mathcal{Y}}^{k}
\end{equation}
where $\hat{\mathcal{Y}}^{k}$ is the predicted clean data at step $k$, obtained from a denoising network $f_{\theta}$ (parameterized by $\theta$):
\begin{equation}
	\hat{\mathcal{Y}}^{k} = f_{\theta}(\mathcal{Y}^{k})
\end{equation}
The network parameters $\theta$ are optimized by minimizing the following mean squared error loss, where the diffusion step $k$ is sampled from a uniform distribution:
\begin{equation}
	\mathcal{L}(\theta)=\mathbb{E}_{k\sim \text{Unif}\{1,...,K\}}[||f_{\theta}(\mathcal{Y}^{k})-\mathcal{Y}^{*}||_{2}^{2}]
\end{equation}
Here, $\mathcal{Y}^{*}$ denotes the ground truth target data.

For the task of time series forecasting (TSF), the diffusion framework is conditioned by introducing historical inputs $\mathcal{Z}$ into the reverse process to guide the generation of future values. Consequently, the reverse denoising distribution is rewritten as:
\begin{equation}
	p(\mathcal{Y}^{k-1}|\mathcal{Y}^{k},\mathcal{Z}) = \mathcal{N}(\mathcal{Y}^{k-1};\ \hat{\mu}(\mathcal{Y}^{k},\mathcal{Z}),\ \sigma_{k}^{2}I)
\end{equation}
The mean $\hat{\mu}(\mathcal{Y}^{k},\mathcal{Z})$ is still calculated according to Equation (3), but the prediction from the denoising network now incorporates the condition $\mathcal{Z}$:
\begin{equation}
	\hat{\mathcal{Y}}^{k} = f_{\theta}(\mathcal{Y}^{k}, \mathcal{Z})
\end{equation}
Therefore, the model is trained by minimizing the following conditional loss function, enabling it to learn probabilistic forecasts for future time series values based on the given historical context:
\begin{equation}
	\mathcal{L}(\theta) = \mathbb{E}_{k\sim \text{Unif}\{1,...,K\}} \left[ ||f_{\theta}(\mathcal{Y}^{k}, \mathcal{Z}) - \mathcal{Y}^{*}||_2^2 \right]
\end{equation}

\subsection{UniDiff Process for TSF}

Leveraging the conditional diffusion framework, the UniDiff process is specifically designed to handle multimodal inputs. In this context, the general conditioning variable $\mathcal{Z}$ encompasses the historical time series $\mathcal{X}$, the complete set of historical and future timestamps $\mathcal{T}_{full} = [\mathcal{T},\mathcal{T}']$, and the associated textual descriptions $\mathcal{D}$. The forward diffusion process remains unchanged, following the standard procedure in Equation (\ref{eq1}), while the reverse process is conditioned on the full spectrum of multimodal information.
During the reverse diffusion process, the denoising network $f_\theta$ learns to iteratively recover the future time series $\mathcal{Y}$ by approximating the following conditional Gaussian distribution:
\begin{equation}
	p(\mathcal{Y}^{k-1}|\mathcal{Y}^{k},\mathcal{X},\mathcal{T}_{full},\mathcal{D}) = \mathcal{N}(\mathcal{Y}^{k-1}; \hat{\mu}_\theta(\mathcal{Y}^k, \mathcal{X}, \mathcal{T}_{full}, \mathcal{D}), \sigma_k^2 I)
\end{equation}
The mean of this distribution, $\hat{\mu}_\theta(\mathcal{Y}^k, \mathcal{X}, \mathcal{T}_{full}, \mathcal{D})$, is calculated according to Equation (3), where the prediction of the clean data, $\hat{\mathcal{Y}}^k$, is now a function of all input modalities:
\begin{equation}
	\hat{\mathcal{Y}}^k = f_\theta(\mathcal{Y}^k, \mathcal{X}, \mathcal{T}_{full}, \mathcal{D})
\end{equation}
The network $f_\theta$ is subsequently trained by minimizing the following multimodal conditional mean squared error loss, which teaches the model to integrate timestamp and textual information to effectively guide the denoising process:
\begin{equation}
	\mathcal{L}(\theta) = \mathbb{E}_{k \sim \text{Unif}\{1,...,K\}} \left[ \big\| f_\theta(\mathcal{Y}^k, \mathcal{X}, \mathcal{T}_{full}, \mathcal{D}) - \mathcal{Y}^* \big\|_2^2 \right]
\end{equation}
Given that the quality and relevance of textual and timestamp information can vary significantly, controlling their influence during the generation process is crucial. To this end, we propose a novel, decoupled Classifier-Free Guidance (CFG) mechanism, tailored for multi-source conditioning. This mechanism allows for independent adjustment of the guidance strength for textual and temporal information at inference time. The final guided prediction $\hat{\mathcal{Y}}^{k}_{\text{guided}}$ is a linear combination of predictions under different conditions:
\begin{equation}
	\hat{\mathcal{Y}}^{k}_{\text{guided}} = \hat{\mathcal{Y}}^{k}_{t, d} + w_t \big(\hat{\mathcal{Y}}^{k}_{t, d} - \hat{\mathcal{Y}}^{k}_{\emptyset_t, d}\big) + w_d \big(\hat{\mathcal{Y}}^{k}_{t, d} - \hat{\mathcal{Y}}^{k}_{t, \emptyset_d}\big)
\end{equation}
Here, $\hat{\mathcal{Y}}^{k}_{t,d} = f_\theta(\mathcal{Y}^k, \mathcal{X}, \mathcal{T}_{full}, \mathcal{D})$ is the fully conditioned prediction, $\hat{\mathcal{Y}}^{k}_{\emptyset_t, d} = f_\theta(\mathcal{Y}^k, \mathcal{X}, \emptyset_t, \mathcal{D})$ is the prediction without the timestamp condition ($\emptyset_t$ being a null token for timestamps), and $\hat{\mathcal{Y}}^{k}_{t, \emptyset_d} = f_\theta(\mathcal{Y}^k, \mathcal{X}, \mathcal{T}_{full}, \emptyset_d)$ is the prediction without the text condition ($\emptyset_d$ being a null token for text). The terms $w_t$ and $w_d$ are scalar hyperparameters that independently adjust the guidance strength of timestamps and text, respectively. These weights are used to steer the generation only during the inference phase.
To enable this decoupled guidance, the $f_\theta$ network must be trained on various combinations of conditional inputs. During training, we randomly drop the timestamp and text conditions and replace them with corresponding null tokens. This process is governed by independent unconditional probabilities $p_{uncond\_\mathcal{T}}$ and $p_{uncond\_\mathcal{D}}$, allowing the model to learn fully conditional, partially conditional, and fully unconditional predictions within a single network.

\subsection{The UniDiff Model}

The UniDiff model is architected around three core components that work in sequence to perform robust forecasting: Modality-specific Encoders to process the diverse inputs, a Unified Multimodal Fusion Module to integrate their representations, and a final Prediction Head to produce the prediction.

\subsubsection{Modality-specific Encoders}

\textbf{a) Temporal Encoder:} To capture local temporal dynamics, we first process the concatenated time series $\mathcal{X}_{full} = [\mathcal{X}, \mathcal{Y}^k]$. This sequence is partitioned into a set of overlapping local patches $\{\mathbf{P}_1, ..., \mathbf{P}_M\}$. Each patch $\mathbf{P}_i$ is flattened into a vector and projected into an embedding space via a shared multi-layer perceptron (MLP) to obtain a patch embedding $\mathbf{z}_i$. The final time series representation is the sequence of these embeddings, i.e., $\mathbf{z} = [\mathbf{z}_1, ..., \mathbf{z}_M]$.

\textbf{b) Timestamp Encoder:} Each timestamp in the set $\mathcal{T}_{full}$ is converted into a feature vector capturing normalized calendar attributes such as the day of the week and the day of the month. This sequence of feature vectors is then processed by a lightweight MLP to obtain the final sequence of timestamp embeddings $\mathbf{t}$.

\textbf{c) Text Encoder:} The text input $\mathcal{D}$ is first tokenized and then encoded by a frozen pre-trained language model (e.g., BERT) to obtain rich semantic embeddings $\mathbf{d}$.

\subsubsection{Unified Multimodal Fusion Module}

Following the initial encoding, the feature embeddings are processed through a stack of $L$ identical fusion layers. The core of our approach lies in the design of these fusion layers, which moves beyond rigid sequential processing to achieve a parallel and synergistic fusion of multimodal representations. Each layer is designed to inject the structural information from the time embeddings $\mathbf{t}$ and the semantic context from the text embeddings $\mathbf{d}$ into the time series embeddings $\mathbf{z}$ through a single, unified operation.

At the core of each fusion layer is a unified cross-attention mechanism. For each layer $l$, the module takes the time series representation from the previous layer, $\mathbf{z}_{l-1}$, as input, while simultaneously integrating information from both the timestamp and text modalities. Specifically, the time series embeddings serve as the \textbf{Query}, while the \textbf{Key} and \textbf{Value} are obtained by concatenating the timestamp and text embeddings. This is formulated as:
\begin{align}
	\mathbf{c} &= \text{Concat}(\lambda \mathbf{t}, \mathbf{d}), \notag \\
	Q = \mathbf{z}_{l-1} W_Q, & \quad  K = \mathbf{c} W_K, \quad V = \mathbf{c} W_V
\end{align}
where $\lambda$ represents the weight used to measure the contribution of timestamp information and $W_Q, W_K, W_V$ are learnable projection matrices. The output of the cross-attention mechanism is:
\begin{equation}
	\text{Attention}(Q, K, V) = \mathrm{softmax}\left(\frac{QK^{T}}{\sqrt{d_k}}\right)V
\end{equation}
The output of this operation is an enhanced time series representation, refined by attending to all available contextual information simultaneously. This is followed by a standard feed-forward network (FFN), with residual connections and layer normalization applied after each sub-layer to ensure training stability. The update rule for the time series representation at layer $l$ is as follows:
\begin{equation}
	\begin{aligned}
		\mathbf{z}_l' &= \mathrm{LayerNorm}(\mathbf{z}_{l-1} + \mathrm{CrossAttention}(\mathbf{z}_{l-1}, \mathbf{c})), \\
		\mathbf{z}_l &= \mathrm{LayerNorm}(\mathbf{z}_l' + \mathrm{FFN}(\mathbf{z}_l'))
	\end{aligned}
\end{equation}
This parallel fusion architecture enables the model to adaptively weigh the importance of timestamp versus textual information at each step. Without a preset processing hierarchy, the model can learn autonomously which modality is more critical for different temporal patterns, leading to more flexible and efficient cross-modal interaction.

\subsubsection{Prediction Head}

After processing through the final unified fusion layer, the model generates a sequence of fused patch embeddings, $\mathbf{z}_L$. The prediction head is designed to transform these rich representations into the final forecast through a parallel, unified, and adaptive fusion mechanism that integrates information from all available modalities.
First, to obtain a holistic representation of the forecast horizon, we apply global average pooling to the sequence of patch embeddings from the final fusion layer:
\begin{equation}
	z_{avg} = \frac{1}{M} \sum_{i=1}^{M} z_{L,i}
\end{equation}
Next, we employ three independent linear prediction heads to generate modality-specific forecasts in parallel. This allows the model to capture the distinct predictive patterns inherent in each data stream:
\begin{equation}
	\hat{y}_z = \mathrm{Linear}_z(z_{avg}), 
	\hat{y}_t = \mathrm{Linear}_t(\mathrm{Pool}(\mathbf{t})), 
	\hat{y}_d = \mathrm{Linear}_d(\mathbf{d})
\end{equation}
Finally, a learnable MLP produces the final output as a dynamically weighted sum of these modality-specific predictions. The fusion weights $\gamma$ are computed adaptively based on the content of the parallel predictions, allowing the network to dynamically lean towards the most informative source at each diffusion step.
\begin{equation}
	\gamma = \mathrm{softmax}(\mathrm{MLP}_{fuse}(\mathrm{Concat}(\hat{y}_z, \hat{y}_t, \hat{y}_d)))
\end{equation}
The final prediction for the current diffusion step, $\hat{y}^k$, is then computed as:
\begin{equation}
	\hat{y}^k = \gamma_z \hat{y}_z + \gamma_t \hat{y}_t + \gamma_d \hat{y}_d
\end{equation}
The resulting $\hat{y}^k$ serves as the denoised estimate in the reverse diffusion process, providing a flexible and adaptive mechanism that fully leverages the information from all multimodal inputs.

\section{Experiments}

\begin{table*}[t]
	\caption{Statistics of benchmark datasets.}
	\label{Multivariate datasets}
	\resizebox{\linewidth}{!}{
		\begin{tabular}{ccccccccc}
			\toprule
			Dataset      & Domain      & Frequency & Lengths & Dim & Split &Stride &Benchmark & Description\\ \midrule
			Agriculture & Retail Broiler Composite & Monthly &496 &1 &7:1:2 &1 &TimeMMD &The record of Retail Broiler Composite between 1983 - Present \\
			Climate & Drought Level & Monthly & 496 & 5 & 7:1:2 & 1 & TimeMMD & The record of Drought Level between 1983 - Present\\
			Economy & International Trade Balance & Monthly & 423 & 3 & 7:1:2 & 1 & TimeMMD & The record of International Trade between 1989 - Present \\
			Energy & Gasoline Prices & Weekly & 1,479 & 9 & 7:1:2 & 1 & TimeMMD & The prices of Gasoline between 1996 - Present \\
			Environment & Air Quaility Index & Daily & 11,102 & 4 & 7:1:2 & 1 & TimeMMD & The indices of Air Quality between 1982 - 2023\\
			Health & Influenza Patients Proportion & Weekly & 1,389 & 11 & 7:1:2 & 1 & TimeMMD & The record of Influenza Patients Proportion between 1997 - Present \\
			Social Good & Unemployment Rate & Monthly & 900 &1 & 7:1:2 &1 & TimeMMD & The Unemployment Rate between 1950 - Present\\
			Traffic & Travel Volume & Monthly & 531 & 1 &7:1:2 & 1 & TimeMMD & The Travel Volume between 1980 - Present\\
			\bottomrule
	\end{tabular}}
\end{table*}

\begin{table*}[!htbp]
	\centering
	\caption{
Comprehensive performance comparison of UniDiff against state-of-the-art baselines across eight real-world multimodal time series datasets. The evaluation is conducted using MSE and MAE, where lower values indicate better forecasting accuracy. For each metric within each dataset, the best-performing model is highlighted in \textcolor{red}{red}.}
	\label{tab: timemmd full}
	\resizebox{\textwidth}{!}{
		\begin{tabular}{cc|cc|cc|cc|cc|cc|cc|cc|cc|cc|cc}
			\toprule
			
			\multicolumn{2}{c|}{\multirow{2}{*}{Models}} & \multicolumn{2}{c|}{UniDiff} & \multicolumn{2}{c|}{Sundial} & \multicolumn{2}{c|}{VisionTS} & \multicolumn{2}{c|}{ROSE} & \multicolumn{2}{c|}{MOIRAI} & \multicolumn{2}{c|}{MCD-TSF} & \multicolumn{2}{c|}{GPT4MTS} & \multicolumn{2}{c|}{CSDI} & \multicolumn{2}{c|}{CALF} & \multicolumn{2}{c}{Time-VLM} \\
			~&~&\multicolumn{2}{c|}{(Ours)} &\multicolumn{2}{c|}{(2025)} & \multicolumn{2}{c|}{(2025)} &\multicolumn{2}{c|}{(2025)} &\multicolumn{2}{c|}{(2024)} & \multicolumn{2}{c|}{(2025)} & \multicolumn{2}{c|}{(2024)} & \multicolumn{2}{c|}{(2021)} & \multicolumn{2}{c|}{(2025)} & \multicolumn{2}{c}{(2025)} \\\midrule
			\multicolumn{2}{c|}{Metrics} & \multicolumn{1}{c}{MSE} & \multicolumn{1}{c|}{MAE} & \multicolumn{1}{c}{MSE} & \multicolumn{1}{c|}{MAE} & \multicolumn{1}{c}{MSE} & \multicolumn{1}{c|}{MAE} & \multicolumn{1}{c}{MSE} & \multicolumn{1}{c|}{MAE} & \multicolumn{1}{c}{MSE} & \multicolumn{1}{c|}{MAE} & \multicolumn{1}{c}{MSE} & \multicolumn{1}{c|}{MAE} & \multicolumn{1}{c}{MSE} & \multicolumn{1}{c|}{MAE} & \multicolumn{1}{c}{MSE} & \multicolumn{1}{c|}{MAE} & \multicolumn{1}{c}{MSE} & \multicolumn{1}{c|}{MAE} & \multicolumn{1}{c}{MSE} & \multicolumn{1}{c}{MAE} \\
			
			\midrule
			\multirow[c]{5}{*}{\rotatebox{90}{Agriculture}} 
			& 6 & \textcolor{red}{\textbf{0.128}} & \textcolor{red}{\textbf{0.234}} & 0.218 & 0.304 & 0.210 & {{0.289}} & 0.220 & 0.299 & {{0.187}} & 0.342 &{{0.137}} &{{0.261}} & 0.161 & 0.257 & {{2.021}} & 0.998 & 0.142 & 0.250 & 0.143 & {{0.245}} \\
			
			& 8 & \textcolor{red}{\textbf{0.191}} & {{0.290}} & 0.319 & 0.364 & 0.266 & {{0.323}} & 0.278 & 0.339 & {{0.245}} & 0.391 & {{0.200}} & 0.318 & 0.207 & 0.288 &{{2.068}} &{{1.029}} & 0.195 & \textcolor{red}{\textbf{0.285}} & 0.215 & 0.287 \\
			
			& 10 & {{0.237}} & {{0.311}} & 0.425 & 0.423 & 0.307 & {{0.348}} & 0.408 & 0.406 & {{0.297}} & 0.423 & {{0.246}} & {{0.339}} &\textcolor{red}{\textbf{0.230}} &\textcolor{red}{\textbf{0.305}} & 2.125 & 1.067 & 0.350 & 0.370 & 0.271 & 0.320 \\
			
			& 12 & \textcolor{red}{\textbf{0.296}} & \textcolor{red}{\textbf{0.341}} & 0.530 & 0.477 & 0.376 & {{0.386}} & 0.474 & 0.443 & {{0.357}} & 0.455 & {{0.305}} &{0.369} & 0.301 & 0.342 &{{2.171}} & {{1.097}} & 0.314 & 0.355 & 0.322 & 0.359 \\ \cmidrule{2-22}
			
			& Avg & \textcolor{red}{\textbf{0.213}} & \textcolor{red}{\textbf{0.294}} & 0.373 & 0.392 & 0.290 & {{0.336}} & 0.345 & 0.372 & {{0.272}} & 0.403 &{{0.222}} &{{0.322}} & 0.225 & {{0.298}} & {{2.096}} & 1.048 & 0.250 & 0.315 & 0.237 & 0.302 \\
			
			\midrule
			\multirow[c]{5}{*}{\rotatebox{90}{Climate}} 
			& 6 &\textcolor{red}{\textbf{1.031}} &\textcolor{red}{\textbf{0.721}} & {{1.180}} & {{0.891}} & 1.316 & 0.932 & 1.488 & 0.993 & 1.624 & 1.016 &{{ 1.588}} & {{0.969}} & 1.199 & {{0.895}} & {{1.278}} & 1.038 & 1.231 & 0.910 & 1.218 & 0.907 \\
			
			& 8 &\textcolor{red}{\textbf{1.049}} &\textcolor{red}{\textbf{0.739}} & {{1.159}} & {{0.885}} & 1.312 & 0.935 & 1.598 & 1.031 & 2.148 & 1.152 & {{1.579}} & {{0.970}} & 1.205 & 0.899 & {{1.262}} & {{1.027}} & 1.227 & 0.905 & 1.181 & 0.914 \\
			
			& 10 &\textcolor{red}{\textbf{1.072}} &\textcolor{red}{\textbf{0.761}} & {{1.141}} & {{0.876}} & 1.302 & 0.928 & 1.401 & 0.967 & 1.983 & 1.112 & {{1.584}} & {{0.969}} & 1.173 & 0.885 & {{1.254}} & 1.022 & 1.508 & 0.990 & 1.179 & {{0.880}} \\
			
			& 12 &\textcolor{red}{\textbf{1.102}} &\textcolor{red}{\textbf{0.768}} & {{1.134}} & {{0.870}} & 1.297 & 0.925 & 1.414 & 0.957 & 1.929 & 1.101 & {{1.590}} & {{0.974}} & {{1.152}} & {{0.876}} & 1.283 & 1.026 & 1.177 & 0.883 & 1.203 & 0.896 \\\cmidrule{2-22}
			
			& Avg &\textcolor{red}{\textbf{1.063}} &\textcolor{red}{\textbf{0.747}} & {{1.154}} & {{0.881}} & 1.307 & 0.930 & 1.475 & 0.987 & 1.921 & 1.095 & {{1.583}} &{{0.971}} & 1.182 & 0.889 & {{1.264}} & {{1.028}} & 1.286 & 0.922 & 1.195 & 0.899 \\
			
			\midrule
			\multirow[c]{5}{*}{\rotatebox{90}{Economy}} 
			& 6 & \textcolor{red}{\textbf{0.015}} & \textcolor{red}{\textbf{0.100}} & {{0.251}} & {{0.401}} & 0.270 & 0.420 & 0.258 & 0.405 & 0.315 & 0.460 &{{0.248}} &{{0.370}} & {{0.016}} & {{0.102}} & 1.492 & 0.966 & 0.178 & 0.334 & 0.024 & 0.125 \\
			
			& 8 & {{0.017}} & \textcolor{red}{\textbf{0.101}} & {{0.277}} & {{0.423}} & 0.296 & 0.440 & 0.300 & 0.450 & 0.431 & 0.526 &{0.248} &{{0.374}} & \textcolor{red}{\textbf{0.016}} & \textcolor{red}{\textbf{0.101}} & 1.492 & 0.967 & 0.200 & 0.353 & 0.023 & 0.121 \\
			
			& 10 & {{0.019}} & \textcolor{red}{\textbf{0.102}} & 0.304 & 0.443 & 0.307 & 0.446 & {{0.286}} & {{0.432}} & 0.432 & 0.528 &{{0.249}} &{0.376} & \textcolor{red}{\textbf{0.018}} & 0.104 & {{1.492}} & {{0.968}} & 0.039 & 0.162 & 0.025 & 0.128 \\
			
			& 12 & 0.019 & \textcolor{red}{\textbf{0.102}} & 0.333 & 0.460 & 0.329 & 0.462 & {{0.310}} & {{0.447}} & 0.440 & 0.535 &{0.249} &0.377 & \textcolor{red}{\textbf{0.017}} & {{0.104}} & 1.492 & 0.970 & 0.232 & 0.378 & 0.024 & 0.126 \\\cmidrule{2-22}
			
			& Avg & 0.018 & \textcolor{red}{\textbf{0.101}} & 0.291 & {{0.432}} & 0.301 & 0.442 & {{0.289}} & 0.433 & 0.405 & 0.512 &0.249 &0.374 & \textcolor{red}{\textbf{0.017}} & {{0.103}} & 1.492 & 0.968 & 0.163 & 0.307 & 0.024 & 0.125 \\
			
			\midrule
			\multirow[c]{5}{*}{\rotatebox{90}{Energy}} 
			& 12 & \textcolor{red}{\textbf{0.102}} & 0.221 & {{0.125}} & {0.242} & 0.173 & 0.313 & 0.268 & 0.401 & 0.183 & 0.309 &0.120 & \textcolor{red}{\textbf{0.212}} & 0.111 & 0.244 & 0.260 & 0.297 & \textcolor{red}{\textbf{0.102}} & {{0.224}} & 0.114 & 0.253 \\
			
			& 24 & \textcolor{red}{\textbf{0.204}} & {{0.331}} & {{0.234}} & 0.345 & 0.264 & 0.395 & 0.363 & 0.469 & 0.290 & 0.396 &0.222 &\textcolor{red}{\textbf{0.322}} & 0.232 & 0.362 & 0.371 & {{0.409}} & {{0.210}} & 0.346 & 0.227 & 0.359 \\
			
			& 36 & \textcolor{red}{\textbf{0.276}} & 0.361 & {{0.318}} & {{0.409}} & 0.346 & 0.454 & 0.413 & 0.497 & 0.367 & 0.449 &0.294 &\textcolor{red}{\textbf{0.352}} & 0.308 & 0.418 & 0.464 & 0.483 & {{0.300}} & 0.420 & 0.309 & {{0.410}} \\
			
			& 48 & \textcolor{red}{\textbf{0.358}} & 0.437 & {{0.410}} & {{0.473}} & 0.434 & 0.516 & 0.501 & 0.549 & 0.457 & 0.515 &0.375 &\textcolor{red}{\textbf{0.431}} & 0.398 & 0.496 & 0.546 & 0.545 & {{0.365}} & {{0.470}} & 0.390 & 0.475 \\\cmidrule{2-22}
			
			& Avg & \textcolor{red}{\textbf{0.235}} & {{0.338}} & {{0.272}} &0.367 & 0.304 & 0.420 & 0.386 & 0.479 & 0.324 & 0.417 &0.253 &\textcolor{red}{\textbf{0.293}} & 0.262 & 0.380 & 0.410 & 0.433 & {{0.244}} & {{0.365}} & 0.260 & 0.374 \\
			
			\midrule
			\multirow[c]{5}{*}{\rotatebox{90}{Environment}} 
			& 48 & {0.284} & \textcolor{red}{\textbf{0.383}} & {{0.330}} & 0.410 & 0.345 & 0.426 & 0.402 & 0.459 & 0.352 &{{0.404}} &\textcolor{red}{\textbf{0.279}} &{0.397} & 0.315 & 0.400 & 0.279 & 0.384 & 0.313 & {{0.382}} & {{0.304}} & 0.387 \\
			
			& 96 & {0.287} & \textcolor{red}{\textbf{0.385}} & {{0.353}} & 0.423 & 0.370 & 0.441 & 0.409 & 0.465 & 0.370 & {{0.415}} &\textcolor{red}{\textbf{0.281}} &{0.398}& 0.340 & 0.401 & 0.306 & 0.397 & 0.335 & {{0.394}} & {{0.327}} & 0.405 \\
			
			& 192 & \textcolor{red}{\textbf{0.273}} & \textcolor{red}{\textbf{0.378}} & {{0.343}} & 0.419 & 0.360 & 0.442 & 0.389 & 0.452 & 0.350 & {{0.402}} &0.279 &0.399& 0.336 & 0.411 & 0.304 & 0.396 & 0.341 & {{0.394}} & {{0.328}} & 0.403 \\
			
			& 336 & 0.272 & \textcolor{red}{\textbf{0.381}} & {{0.317}} & 0.411 & 0.340 & 0.436 & 0.369 & 0.447 & 0.332 & {{0.390}} &\textcolor{red}{\textbf{0.261}} &0.393 & {{0.299}} & 0.390 & 0.274 & 0.386 & 0.312 & {{0.377}} & 0.320 & 0.395 \\\cmidrule{2-22}
			
			& Avg & 0.279 & \textcolor{red}{\textbf{0.382}} & {{0.336}} & 0.416 & 0.354 & 0.436 & 0.392 & 0.456 & 0.351 & {{0.403}} &\textcolor{red}{\textbf{0.275}} &0.397 & 0.323 & 0.400 & 0.291 & 0.391 & 0.325 & {{0.387}} & {{0.319}} & 0.397 \\
			
			\midrule
			\multirow[c]{5}{*}{\rotatebox{90}{Health}} 
			& 12 & {1.002} & 0.642 & {{1.531}} & {{0.810}} & 2.012 & 1.093 & 2.737 & 1.250 & 2.230 & 1.114 & 1.145 & 0.676 & 0.985 & 0.658 & 1.228 & {{0.702}} & \textcolor{red}{\textbf{0.964}} &\textcolor{red}{\textbf{0.609}} &1.198 & 0.727 \\
			
			& 24 & \textcolor{red}{\textbf{1.342}} & 0.798 & {{2.075}} & {{1.019}} & 2.594 & 1.266 & 2.589 & 1.189 & 2.895 & 1.284 & {{1.485}} & 0.831 & 1.513 & 0.802 &1.636 & {{0.849}} & 1.451 &\textcolor{red}{\textbf{0.749}} & {{1.491}} & 0.839 \\
			
			& 36 & \textcolor{red}{\textbf{1.477}} & \textcolor{red}{\textbf{0.820}} & {{2.122}} & {{1.058}} & 2.686 & 1.291 & 2.629 & 1.210 & 2.924 & 1.289 &1.620 &0.853 & 1.601 & 0.846 & {{1.852}} & {{0.917}} & 1.713 & 0.851 & 1.567 & 0.865 \\
			
			& 48 & \textcolor{red}{\textbf{1.589}} & \textcolor{red}{\textbf{0.849}} & {{2.153}} & {{1.081}} & 2.454 & 1.236 & 2.436 & 1.154 & 2.895 & 1.276 &1.732 &0.882 & 1.757 & 0.889 & {{2.022}} & {{0.962}} & 1.836 & 0.889 & 1.702 & 0.907 \\\cmidrule{2-22}
			
			& Avg & \textcolor{red}{\textbf{1.353}} & \textcolor{red}{\textbf{0.778}} & {{1.970}} & {{0.992}} & 2.436 & 1.221 & 2.598 & 1.201 & 2.736 & 1.241 &1.496 & 0.811 & 1.464 & 0.799 & {{1.685}} &{0.857} & 1.491 & {{0.775}} & 1.489 & 0.834 \\\midrule
			
			\multirow[c]{5}{*}{\rotatebox{90}{Social Good}} 
			& 6 & {0.729} & {0.379} & {{0.861}} & {{0.487}} & 0.957 & 0.543 & 0.939 & 0.499 & 0.966 & 0.522 &{0.910} & 0.502 & \textcolor{red}{\textbf{0.718}} & 0.378 & 0.947 & {{0.515}} & 0.782 &\textcolor{red}{\textbf{0.360}} & 0.732 & 0.379 \\
			
			& 8 & 0.824 & {0.507} & {{0.994}} & {{0.549}} & 1.106 & 0.605 & 1.168 & 0.588 & 1.532 & 0.653 &{1.005} & 0.536 & 0.942 & 0.505 & 1.069 & {{0.554}} & 0.874 &\textcolor{red}{\textbf{0.386}} & \textcolor{red}{\textbf{0.822}} & 0.427 \\
			
			& 10 & \textcolor{red}{\textbf{0.890}} & 0.448 & {{1.100}} & {{0.604}} & 1.164 & 0.636 & 1.187 & 0.595 & 1.551 & 0.691 &1.071 & 0.607 & 0.929 & {0.446} & 1.185 & 0.604 & 0.976 &\textcolor{red}{\textbf{0.420}} & {{0.916}} & 0.465 \\
			
			& 12 & \textcolor{red}{\textbf{0.971}} & {0.472} & {{1.187}} & {{0.651}} & 1.278 & 0.688 & 1.272 & 0.642 & 1.671 & 0.736 &1.152 & 0.629 & 1.093 &{{0.470}} & 1.247 & 0.619 & {{0.991}} &\textcolor{red}{\textbf{0.439}} & 1.005 & 0.505 \\\cmidrule{2-22}
			
			& Avg & \textcolor{red}{\textbf{0.854}} & {0.452} & {{1.036}} & {{0.573}} & 1.126 & 0.618 & 1.141 & 0.581 & 1.430 & 0.651 &\textcolor{red}{\textbf{1.035}} & 0.569 & 0.920 & 0.450 & 1.112 & {{0.573}} & 0.906 &\textcolor{red}{\textbf{0.401}} & {{0.868}} & 0.444 \\
			
			\midrule
			\multirow[c]{5}{*}{\rotatebox{90}{Traffic}} 
			& 6 & 0.179 & {0.284} & {{0.273}} & {{0.410}} & 0.275 & 0.411 & 0.331 & 0.449 & 0.349 & 0.448 &0.186 & 0.297 & 0.192 & 0.264 & {{0.202}} &0.326 & \textcolor{red}{\textbf{0.174}} & \textcolor{red}{\textbf{0.243}} & 0.210 & 0.316 \\
			
			& 8 & 0.185 & 0.285 & {{0.275}} & {{0.408}} & 0.282 & 0.410 & 0.365 & 0.455 & 0.461 & 0.499 &0.191 & 0.289 & 0.195 & 0.256 & 0.216 & {{0.342}} & \textcolor{red}{\textbf{0.176}} & \textcolor{red}{\textbf{0.232}} & 0.212 &0.313 \\
			
			& 10 & \textcolor{red}{\textbf{0.190}} & \textcolor{red}{\textbf{0.288}} & {{0.270}} & {{0.403}} & 0.286 & 0.406 & 0.326 & 0.443 & 0.414 & 0.466 &0.196 & 0.292 & 0.204 & {{0.257}} & {{0.223}} & 0.343 & 0.345 & 0.454 & 0.222 &0.328 \\
			
			& 12 & \textcolor{red}{\textbf{0.195}} & 0.293 & {{0.268}} & {{0.401}} & 0.282 & 0.402 & 0.342 & 0.458 & 0.400 & 0.458 &{0.201} & 0.301 & 0.218 & 0.268 & {{0.227}} &0.342 & 0.193 & \textcolor{red}{\textbf{0.243}} & 0.222 & 0.322 \\\cmidrule{2-22}
			
			& Avg & \textcolor{red}{\textbf{0.187}} & \textcolor{red}{\textbf{0.288}} & {{0.271}} & {{0.405}} & 0.281 & 0.407 & 0.341 & 0.451 & 0.406 & 0.468 &{0.193} & 0.295 & 0.203 & {{0.261}} & {{0.215}} &0.338 & 0.222 & 0.293 & 0.216 & 0.319 \\\midrule
			
			\multicolumn{2}{c|}{1\textsuperscript{st} Count} & \textcolor{red}{\textbf{26}} & \textcolor{red}{\textbf{23}} & {{0}} & {{0}} & 0 & 0 & 0 & 0 & 0 & 0 &5 &5 & 6 & 2 & 0 & 0 & 4 & 11 & 1 & 0 \\
			\bottomrule
		\end{tabular}
	}
\end{table*}

\begin{table*}[hb]
	\centering

	\caption{Results of the ablation study, designed to quantify the contribution of each core component of the UniDiff framework. We evaluate the full model against six ablated variants across all eight benchmark datasets using both MSE and MAE metrics.}
	\resizebox{1\linewidth}{!}{
		\begin{tabular}{c|cc|cc|cc|cc|cc|cc|cc|cc}
			\toprule
			\multirow{2}{*}{Variants}
			& \multicolumn{2}{c|}{Agriculture}
			& \multicolumn{2}{c|}{Climate}
			& \multicolumn{2}{c|}{Economy}
			& \multicolumn{2}{c|}{Energy}
			& \multicolumn{2}{c|}{Environment}
			& \multicolumn{2}{c|}{Health}
			& \multicolumn{2}{c|}{Social Good}
			& \multicolumn{2}{c}{Traffic} \\
			\cmidrule(lr){2-3}
			\cmidrule(lr){4-5}
			\cmidrule(lr){6-7}
			\cmidrule(lr){8-9}
			\cmidrule(lr){10-11}
			\cmidrule(lr){12-13}
			\cmidrule(lr){14-15}
			\cmidrule(lr){16-17}
			& \makecell{MSE} & \makecell{MAE}
			& \makecell{MSE} & \makecell{MAE}
			& \makecell{MSE} & \makecell{MAE}
			& \makecell{MSE} & \makecell{MAE}
			& \makecell{MSE} & \makecell{MAE}
			& \makecell{MSE} & \makecell{MAE}
			& \makecell{MSE} & \makecell{MAE}
			& \makecell{MSE} & \makecell{MAE} \\
			\midrule
			\cellcolor{lightgray!30}UniDiff (Full Model) & \cellcolor{lightgray!30}0.213 & \cellcolor{lightgray!30}0.294 & \cellcolor{lightgray!30}1.063 & \cellcolor{lightgray!30}0.747 & \cellcolor{lightgray!30}0.018 & \cellcolor{lightgray!30}0.101 & \cellcolor{lightgray!30}0.235 & \cellcolor{lightgray!30}0.338 & \cellcolor{lightgray!30}0.279 & \cellcolor{lightgray!30}0.382 & \cellcolor{lightgray!30}1.353 & \cellcolor{lightgray!30}0.778 & \cellcolor{lightgray!30}0.854 & \cellcolor{lightgray!30}0.452 & \cellcolor{lightgray!30}0.187 & \cellcolor{lightgray!30}0.288 \\
			w/o Text & 0.347 & 0.394 & 1.457 & 0.996 & 0.398 & 0.512 & 0.323 & 0.425 & 0.312 & 0.410 & 1.972 & 0.994 & 1.112 & 0.573 & 0.235 & 0.360 \\
			w/o Timestamp & 0.289 & 0.335 & 1.301 & 0.928 & 0.305 & 0.443 & 0.305 & 0.418 & 0.291 & 0.391 & 1.685 & 0.857 & 0.977 & 0.691 & 0.215 & 0.338 \\
			w/o Both (Unimodal) & 2.096 & 1.048 & 1.264 & 1.028 & 1.492 & 0.968 & 0.410 & 0.433 & 0.355 & 0.437 & 2.463 & 1.222 & 1.216 & 0.961 & 0.289 & 0.470 \\
			Sequential Fusion & 0.222 & 0.322 & 1.583 & 0.971 & 0.249 & 0.374 & 0.253 & 0.293 & 0.275 & 0.397 & 1.496 & 0.811 & 1.035 & 0.569 & 0.193 & 0.295 \\
			Simple Fusion & 0.290 & 0.336 & 1.287 & 0.991 & 0.415 & 0.676 & 0.306 & 0.426 & 0.325 & 0.441 & 1.646 & 0.997 & 1.148 & 0.600 & 0.213 & 0.361 \\
			Coupled CFG & 0.238 & 0.312 & 1.195 & 0.889 & 0.349 & 0.488 & 0.277 & 0.394 & 0.288 & 0.390 & 1.468 & 0.877 & 1.030 & 0.588 & 0.206 & 0.309 \\
			\bottomrule
		\end{tabular}
	}

	\label{table2}
\end{table*}

In this section, we conduct extensive experiments on eight real-world multimodal time series benchmark datasets to evaluate the performance of the proposed Unidiff framework. Additionally, we carry out comprehensive ablation studies and visualization analyses on the various components of the Unidiff framework.

\subsection{Experiments Settings}

\subsubsection{Datasets and Evaluation Metrics}

The experiments in this study are conducted using the Time-MMD \cite{liu2024time}, a comprehensive and pioneering multi-domain, multimodal time series dataset. 
This dataset is distinguished by its extensive coverage, and we use eight distinct real-world datasets: Agriculture, Climate, Economy, Energy, Environment, Health, Social Good, Security, and Traffic.
A key contribution of Time-MMD is its meticulous integration of numerical time series with corresponding textual data, addressing a significant gap in the resources available for multimodal time series analysis. The detailed information of datasets is in Table \ref{Multivariate datasets}.

Time-MMD falls under deterministic forecasting, and we employ the MSE and MAE as evaluation metrics, according to the following formulas:
\begin{equation}
	MAE = \text{mean}\left(|\hat{y} - y^*|\right)
\end{equation}
\begin{equation}
	MSE = \sqrt{\text{mean}\left(|\hat{y} - y^*|\right)}
\end{equation}
where $\hat{y}$ and $y^*$ denote the predicted and true values, respectively.

\subsubsection{Baselines}

We compare our proposed model with state-of-the-art methods from four categories: (1) models that adapt existing Large Language and Vision-Language Models, including CALF \cite{liu2025calf}, Time-VLM \cite{zhong2025timevlm}, and GPT4MTS \cite{jia2024gpt4mts}; 
(2) Transformer-based foundation models built specifically for time series, such as Sundial \cite{liu2025sundial}, ROSE \cite{wang2025towards} and MOIRAI \cite{woo2024unified}; 
(3) models that apply computer vision techniques, represented by VisionTS \cite{chen2025visionts}; and (4) generative diffusion models, such as MCD-TSF \cite{su2025multimodal} and CSDI \cite{tashiro2021csdi}.

\begin{itemize}
	\item \textbf{Sundial}\cite{liu2025sundial} is a family of native time series foundation models that introduces a flow-matching based TimeFlow loss function, enabling direct generative pretraining on continuous-valued time series without the need for discretization or tokenization.
	
	\item \textbf{VisionTS}\cite{chen2025visionts} reconstructs time series forecasting as an image restoration task, treating time series data as images so that a pretrained visual masked autoencoder can perform zero-shot prediction.
	
	\item \textbf{ROSE}\cite{wang2025towards} is a universal time series forecasting model that learns unified representations from heterogeneous data via Decomposed Frequency Learning, while utilizing a novel Time Series Register to capture domain-specific features for adaptive transfer.
	
	\item \textbf{CALF}\cite{liu2025calf} is a cross-modal fine-tuning framework that systematically reduces distributional discrepancies between time data and textual data at multiple levels (input, feature, and output), thereby aligning large language models for use in time series forecasting.
	
	\item \textbf{Time-VLM}\cite{zhong2025timevlm} is a multimodal framework that leverages pretrained Vision-Language Models to enhance prediction by combining rich temporal features with both visual and textual representations of time series data.
	
	\item \textbf{MOIRAI}\cite{woo2024unified} is a general-purpose Transformer model for time series forecasting, which introduces architectural augmentations to handle varying frequencies, arbitrary numbers of variables, and diverse data distributions, and is designed for large-scale pretraining on heterogeneous data.
	
	\item \textbf{GPT4MTS}\cite{jia2024gpt4mts} is a prompt-based framework that prepends trainable soft prompts derived from text information to numerical time series embeddings, thus enabling large language models for multimodal time series forecasting.
	
	\item \textbf{MCD-TSF}\cite{su2025multimodal} is a multimodal conditional diffusion model that uses timestamps and textual descriptions as supplementary guiding information to control the generative denoising process, thereby enhancing time series prediction performance.
	
	\item \textbf{CSDI}\cite{tashiro2021csdi} is a conditional score-based diffusion model designed for probabilistic time series imputation, explicitly learning the conditional distribution of missing values based on observed data.
\end{itemize}

\subsubsection{Implement Details}

For our implementation, we follow established preprocessing standards by configuring look-back and forecast horizons based on the data's temporal frequency (monthly, weekly, and daily) and normalizing all numerical series to zero mean and unit variance. The UniDiff model processes the time series by tokenizing it into patches of length $P=16$ with a stride of $S=8$, which are then embedded using an MLP. Timestamps are converted into normalized calendar features, while textual input is generated by concatenating reports from the 36 intervals preceding the forecast period. Our model architecture employs a frozen \texttt{BERT-base} for text encoding, a stack of $L=6$ unified fusion layers with a hidden dimension of $d_{\text{model}}=512$, and 8-head attention. The diffusion process utilizes a quadratic noise schedule with $K=200$ steps and a DDIM sampler for inference. The model is trained using the Adam optimizer, and for our decoupled classifier-free guidance, we set default guidance strengths of $w_t=0.5$ and $w_d=0.8$, with unconditional training probabilities $p_{\text{uncond\_T}}=0.1$ and $p_{\text{uncond\_D}}=0.1$. To ensure statistical reliability, all reported results are the average of three separate runs with different random seeds.

\subsection{Main Results}
To validate the overall effectiveness of our proposed framework, we conduct a comprehensive comparison against a range of state-of-the-art models, with the detailed results presented in Table \ref{tab: timemmd full}. The empirical evidence shows that UniDiff consistently establishes a new state-of-the-art in multimodal time series forecasting, achieving the best overall performance. Our model demonstrates remarkable consistency by securing the top performance on a majority of the eight datasets, including Agriculture, Energy, and Traffic. The significant performance improvement over MCD-TSF, the previous leading multimodal diffusion model, underscores the superiority of UniDiff's unified parallel fusion and decoupled guidance mechanism compared to earlier sequential fusion strategies. Furthermore, while strong LLM-based methods like GPT4MTS exhibit competitive performance on specific domains such as Economy, UniDiff displays greater overall robustness and generalizability across the diverse set of benchmarks. The substantial performance gap between UniDiff and the unimodal diffusion model CSDI further validates the critical importance of effectively integrating and leveraging auxiliary textual and temporal information, a core strength of our framework. These results collectively affirm that the UniDiff architecture provides a more powerful and flexible approach to modeling complex multimodal time series, setting a new performance benchmark for the task.

\begin{figure*}[hbpt]
	\centering
	\includegraphics[width=1\linewidth]{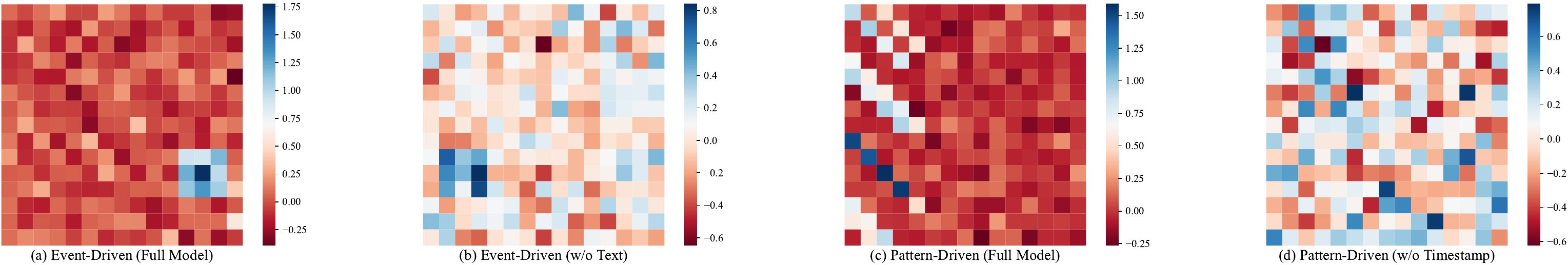}

	\caption{Visualization of cross-attention weights in the Unified Fusion Module. The heatmaps contrast the attention focus of the full UniDiff model against ablated variants. (a, b) Event-Driven Scenario: The full model (a) distinctively attends to textual cues to capture unexpected flu outbreaks, whereas the `w/o Text' variant (b) lacks this focal point. (c, d) Pattern-Driven Scenario: The full model (c) leverages timestamp embeddings to capture seasonal periodicity, while the `w/o Timestamp' variant (d) exhibits a scattered attention pattern, failing to recognize the temporal structure.}
	\label{figure_relitu}

\end{figure*}

\subsection{Ablation Study}
To thoroughly investigate the source of UniDiff's performance and validate our architectural choices, we conduct a series of ablation experiments where we systematically remove or replace key components of the model. We designed several variants to test the impact of each contribution. To assess the importance of multimodal inputs, we test variants without the text modality (\texttt{w/o Text}), without the timestamp modality (\texttt{w/o Timestamp}), and a purely unimodal version that uses neither (\texttt{w/o Both}). To validate our fusion architecture, we include a \texttt{Sequential Fusion} variant that replaces our parallel module with a step-by-step process akin to prior work, and a \texttt{Simple Fusion} variant that substitutes the cross-attention mechanism with basic feature concatenation. Finally, to demonstrate the benefit of our novel guidance mechanism, we test a \texttt{Coupled CFG} variant that reverts our novel decoupled guidance to a standard single-weight mechanism, to test the value of independent modal control.

\subsubsection{Quantitative Analysis}
The results, summarized in Table~\ref{table2}, unequivocally demonstrate that the complete UniDiff model achieves the best performance, with every ablation leading to a degradation in forecasting accuracy. The most significant performance drop is observed in the unimodal variant (\texttt{w/o Both}), which increases the average MSE by over 135\%, confirming the fundamental importance of leveraging multimodal data. Notably, removing the text modality (\texttt{w/o Text}) results in a more substantial performance loss than removing timestamps (\texttt{w/o Timestamp}), highlighting the critical role that semantic context plays in accurately forecasting events within these datasets. The superiority of our unified parallel fusion module is also clearly validated. The \texttt{Sequential Fusion} variant yields a considerable increase in error, while the \texttt{Simple Fusion} variant performs even worse, confirming that our sophisticated cross-attention mechanism is crucial for effectively capturing complex inter-modal dynamics. Finally, the experiment with a \texttt{Coupled CFG} shows a distinct decline in performance compared to our full model. This result validates the effectiveness of our decoupled guidance strategy, proving that providing independent control over the influence of each modality is essential for robust and accurate prediction.

\subsubsection{Qualitative Analysis}
To qualitatively validate our model's adaptive fusion mechanism, we visualize the attention heatmaps of the fusion module using a representative case from the Health dataset (Figure \ref{figure_relitu}). We analyze two distinct forecasting drivers:
1) Event-Driven Prediction (Figure \ref{figure_relitu} a \& b): This scenario involves a sudden, non-seasonal flu outbreak triggered by external factors. As shown in Figure \ref{figure_relitu}(a), the full UniDiff model exhibits a concentrated attention distribution, indicating it successfully captures semantic cues from news reports (e.g., "flu cases surged") to predict the anomaly. In contrast, Figure \ref{figure_relitu}(b) shows that removing the text modality results in a dispersed attention pattern, limiting the model's ability to foresee such event-driven spikes.
2) Pattern-Driven Prediction (Figure \ref{figure_relitu} c \& d): This scenario reflects the regular seasonal cycle of influenza. Figure \ref{figure_relitu}(c) demonstrates that the full model effectively attends to timestamp features (e.g., month of the year) to model periodic trends. However, when timestamps are removed in Figure \ref{figure_relitu}(d), the model loses this structural guidance, leading to an incoherent attention map. These comparisons confirm that UniDiff adaptively shifts its attention between textual and temporal modalities depending on the forecasting context.

\begin{figure}[ht]
	\begin{subfigure}[b]{0.49\linewidth}
		\includegraphics[width=\linewidth]{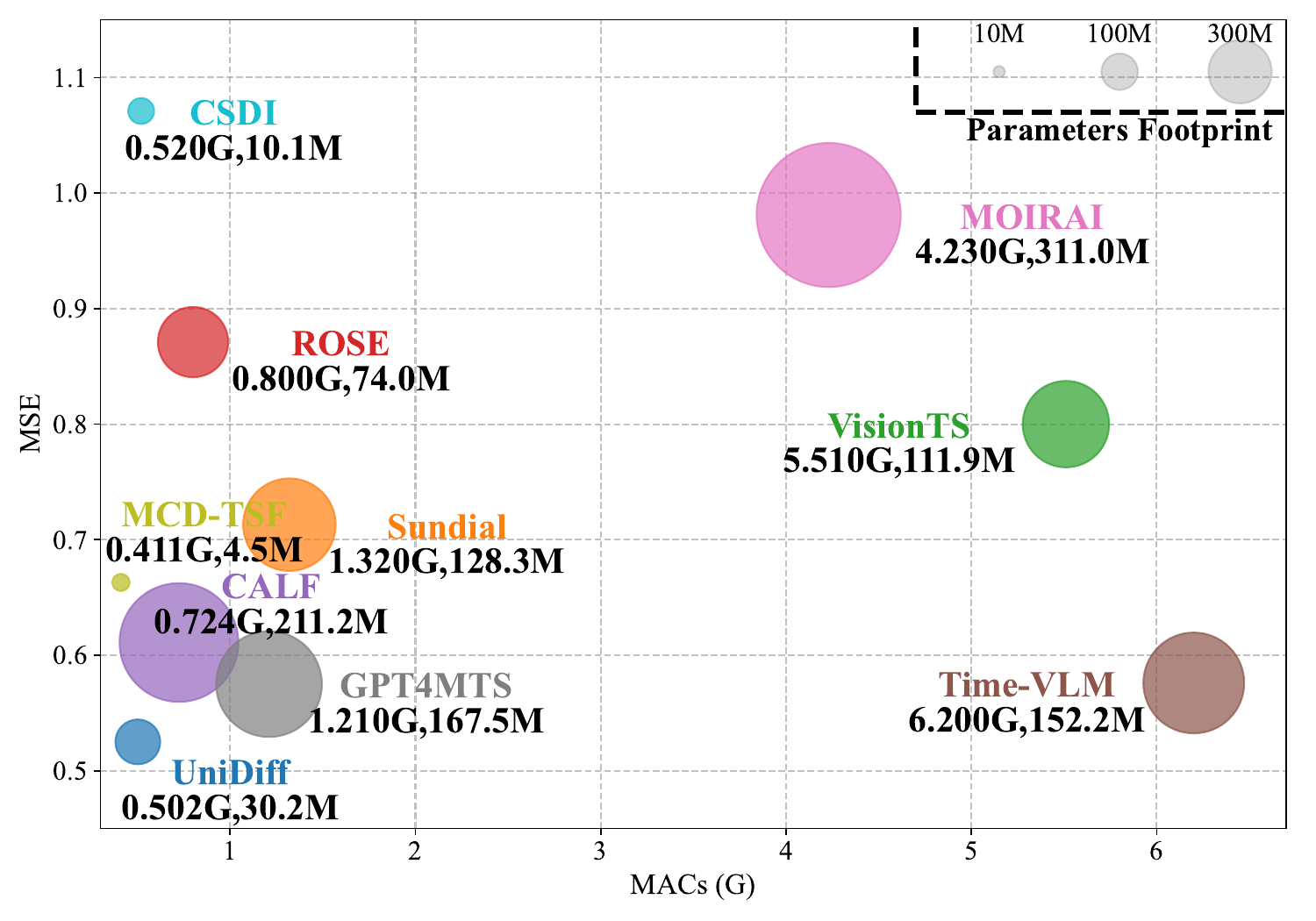}
		
	\end{subfigure}
	\hfill
	\begin{subfigure}[b]{0.496\linewidth}
		\includegraphics[width=\linewidth]{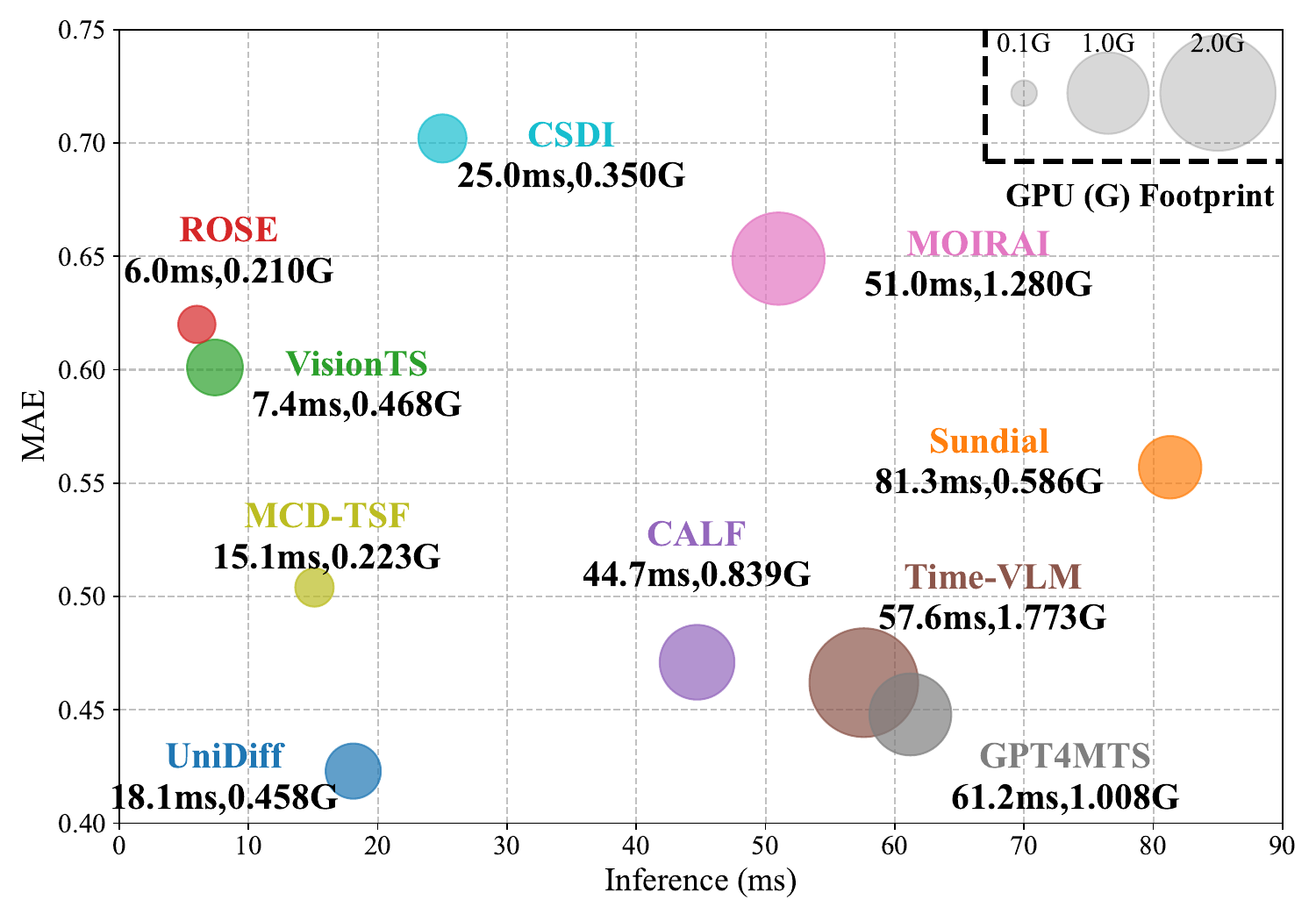}		
	\end{subfigure}
	\caption{Efficiency comparison, evaluated with prediction horizon of 336 and batch size of 1. Left: MSE and computational complexity (MACs), with bubble size indicating parameter count. Right: MAE and inference speed (ms), with bubble size representing peak GPU memory usage.}
	\label{figure_efficiency}
\end{figure}

\subsection{Additional Experiments}
\subsubsection{Efficiency Analysis}

To assess the practical viability of our proposed framework, we conduct a thorough efficiency analysis, evaluating UniDiff against key baselines in terms of computational complexity, inference speed, and memory usage. The results, visualized in Figure \ref{figure_efficiency}, demonstrate that UniDiff strikes an exceptional balance between state-of-the-art accuracy and practical efficiency. The left panel, which positions models based on their MSE and computational load in MACs, shows that UniDiff occupies the most favorable position, achieving the lowest MSE with a modest computational footprint. While models like CSDI and ROSE are lighter, they suffer from significantly higher prediction errors, whereas powerful baselines like MOIRAI and Time-VLM require substantially more computation yet fail to match UniDiff's accuracy. The right panel reinforces this by examining inference speed and memory consumption, where UniDiff again delivers the lowest MAE with a highly competitive inference time. Although some models like MCD-TSF are marginally faster, their accuracy is lower, while others like Sundial and GPT4MTS are considerably slower and more memory-intensive without a corresponding performance benefit. This comprehensive analysis confirms that UniDiff's architecture is not only highly effective but also efficient, making it well-suited for practical applications where both high accuracy and timely predictions are essential.

\begin{figure*}[ht]
	\centering
	\includegraphics[width=1\linewidth]{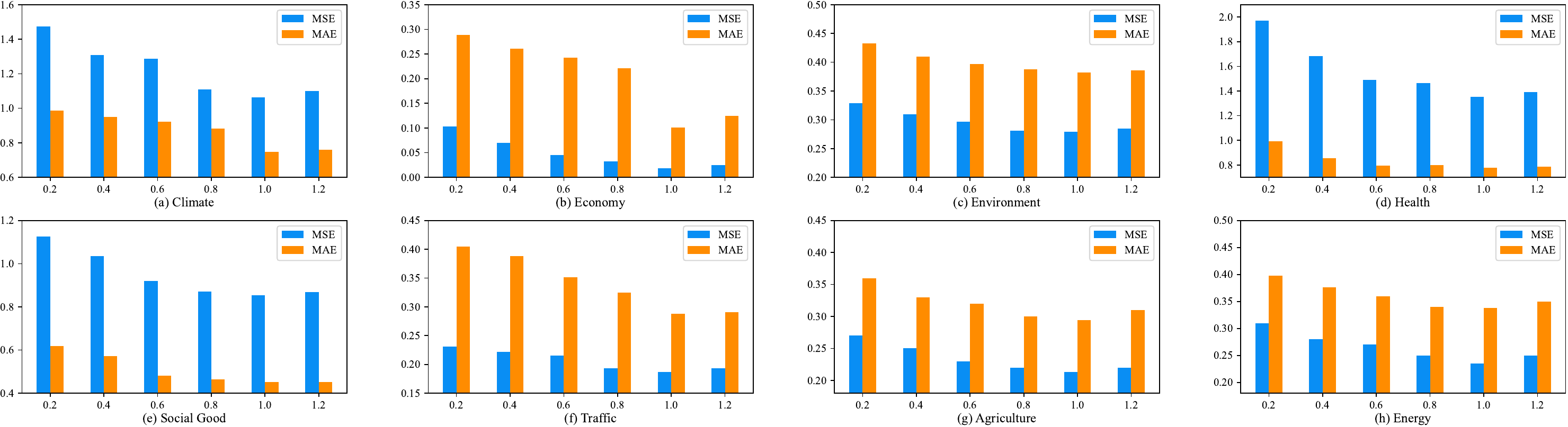}
	\caption{Performance of UniDiff on eight datasets when configured with different values for the timestamp weight $\lambda$.}
	\label{figure_lambda}
\end{figure*}

\subsubsection{Analysis of Timestamp Weight $\lambda$}

To explore the effect of the timestamp modality, we analyze the model's performance under different configurations of the timestamp weight $\lambda$. Figure~\ref{figure_lambda} presents the results on the eight datasets, showing a consistent trend where higher values of $\lambda$ lead to improved performance. 
Specifically, as the weight increases from 0.2 to 1.0, both MSE and MAE scores steadily decrease, indicating that giving more prominence to timestamp information enhances the model's forecasting accuracy. 
The optimal performance in most domains is achieved when $\lambda$ is set to 1.0. 
A slight increase in error is observed when the weight is further increased to 1.2, which suggests that while timestamps are beneficial, an excessive weight can slightly diminish performance. This analysis confirms the effectiveness of incorporating timestamp information and validates our choice of balancing its contribution within the fusion process.

\subsubsection{Analysis of CFG Weight}

\begin{figure*}[hb]
	\centering
	\includegraphics[width=1\linewidth]{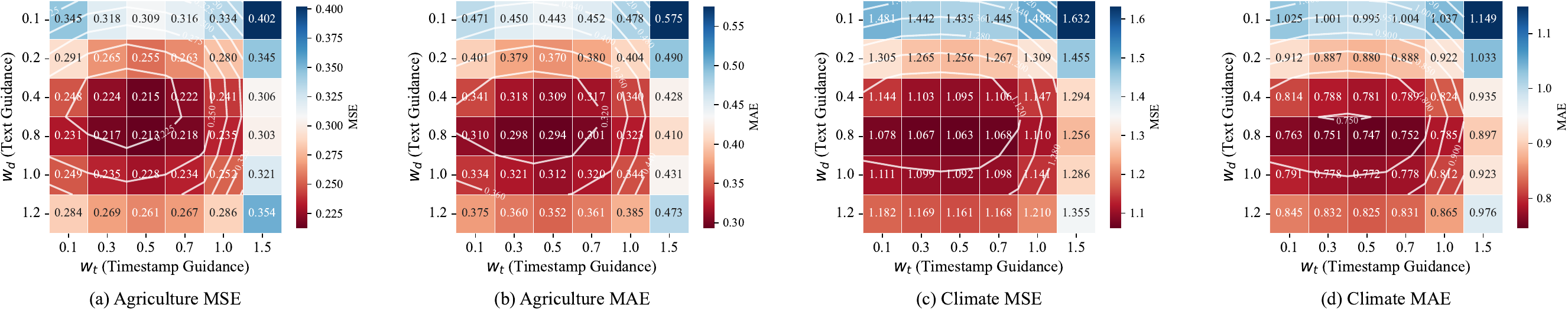}
	\caption{Hyperparameter sensitivity analysis of the decoupled guidance weights ($w_t$ and $w_d$) on the Agriculture \& Climate dataset. Darker colors signify better performance.}
	\label{figure_w}

\end{figure*}

To investigate the impact of our decoupled classifier-free guidance mechanism, we perform a sensitivity analysis on the guidance weights for timestamps ($w_t$) and text ($w_d$). Figure \ref{figure_w} presents the results on the Agriculture and Climate dataset, where the heatmap color visualizes the MSE and the annotated values represent the MAE for each combination of weights. 
The results clearly indicate that the model's performance is sensitive to these parameters, validating the importance of guidance. Performance is weakest when both guidance strengths are set to low values, confirming that conditioning is crucial for accurate forecasting. We observe that the optimal performance, characterized by the lowest MSE and MAE, is concentrated in a region where the timestamp guidance is moderate ($w_t$ around 0.5) and the text guidance is slightly stronger ($w_d$ between 0.8 and 1.0), which supports our choice of default values. The analysis also reveals that excessively high guidance strengths, particularly for the text modality, can lead to a slight degradation in performance, likely because the model starts to overfit to potentially noisy textual cues. Nevertheless, the model demonstrates robustness, as performance remains stable across a reasonable range of values around the optimal configuration, highlighting the effective and controllable nature of our decoupled guidance mechanism.

\subsubsection{Analysis of $p_{uncond}$}

\begin{figure*}[ht]
	\centering
	\includegraphics[width=1\linewidth]{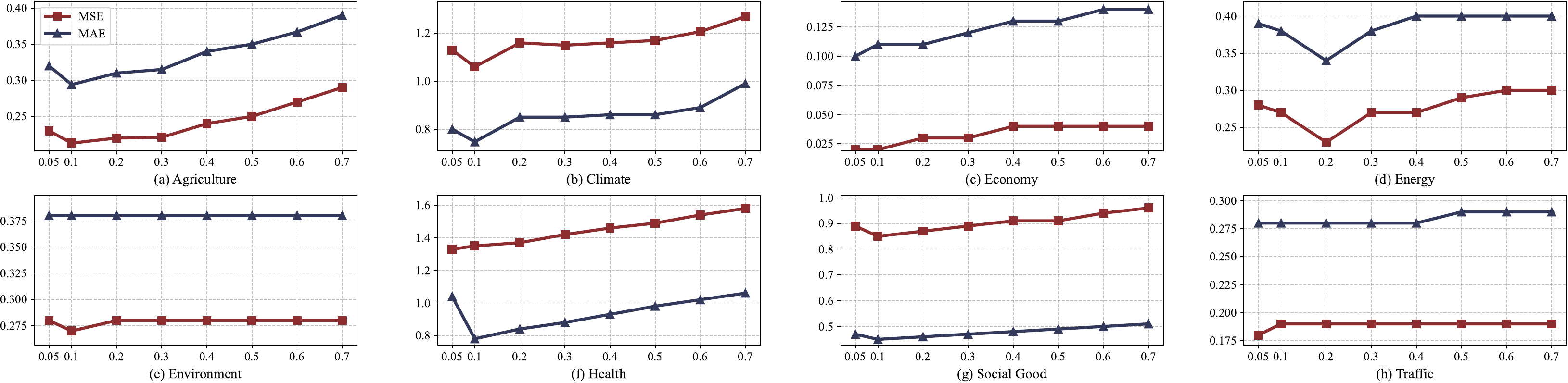}
	\caption{The effect of the unconditional training probability ($p_{uncond}$) on model performance across eight datasets.}
	\label{figure_p}

\end{figure*}

We analyze the impact of the unconditional training probabilities, $p_{uncond\_T}$ and $p_{uncond\_D}$, which are key hyperparameters for our decoupled guidance mechanism. To simplify the analysis and unify control over dropping text and timestamp guidance in this experiment, we set both parameters to an identical value, denoted as the unified parameter $p_{uncond}$. Figure \ref{figure_p} shows the results for eight datasets, revealing a consistent trend in model performance. 
Specifically, the lowest MSE and MAE are achieved when $p_{uncond}$ is set to a low value of 0.1. As this probability increases, the forecasting error for both metrics steadily rises. This is an intuitive result; while a small amount of unconditional training is necessary for our decoupled guidance mechanism to function, excessively high probabilities mean the model spends too much time learning to denoise without the crucial context provided by the modalities, thus degrading its conditional forecasting accuracy. 

\subsubsection{Case Study}

\begin{figure}[hb]
	\begin{subfigure}[b]{0.32\linewidth}
		\includegraphics[width=\linewidth]{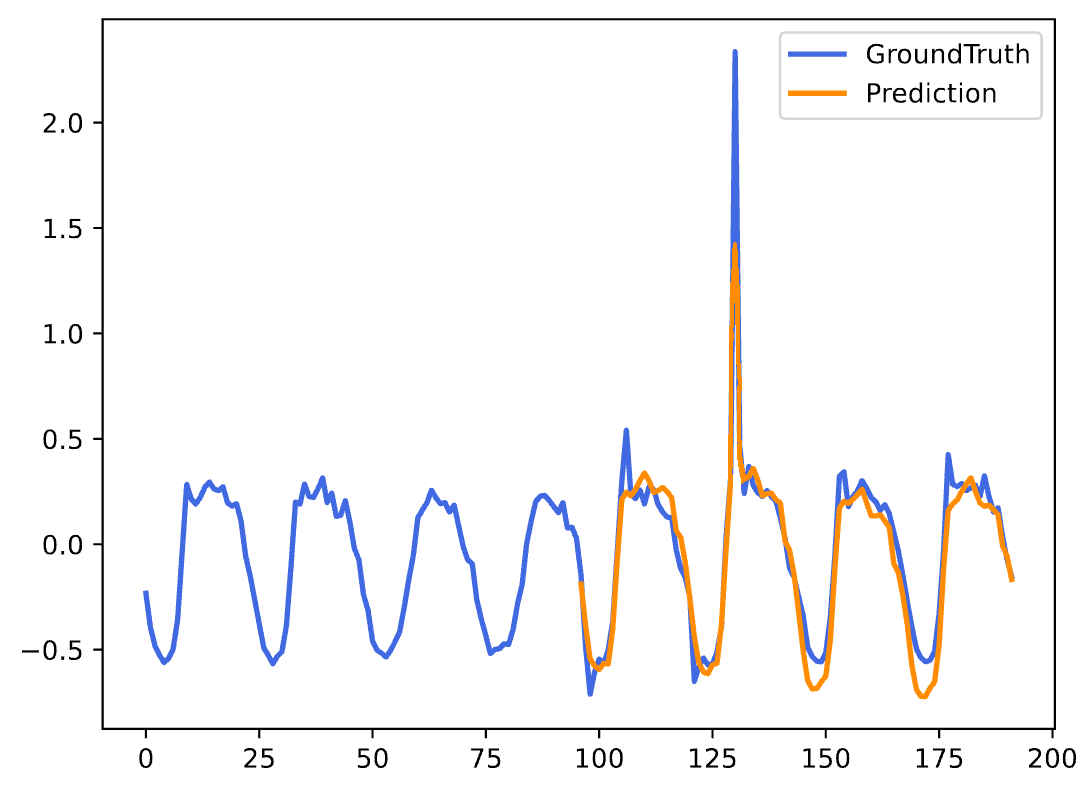}
		\caption{UniDiff}
	\end{subfigure}
	\hfill
	\begin{subfigure}[b]{0.32\linewidth}
		\includegraphics[width=\linewidth]{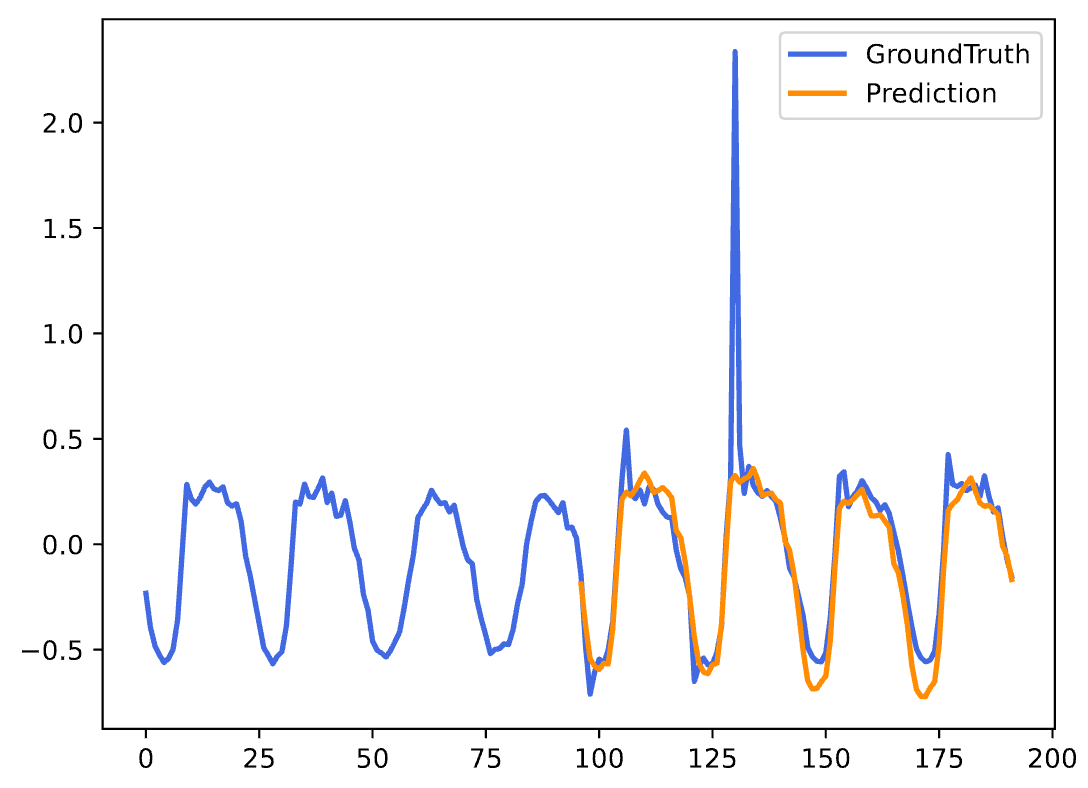}		
		\caption{MCD-TSF}
	\end{subfigure}
	\begin{subfigure}[b]{0.32\linewidth}
		\includegraphics[width=\linewidth]{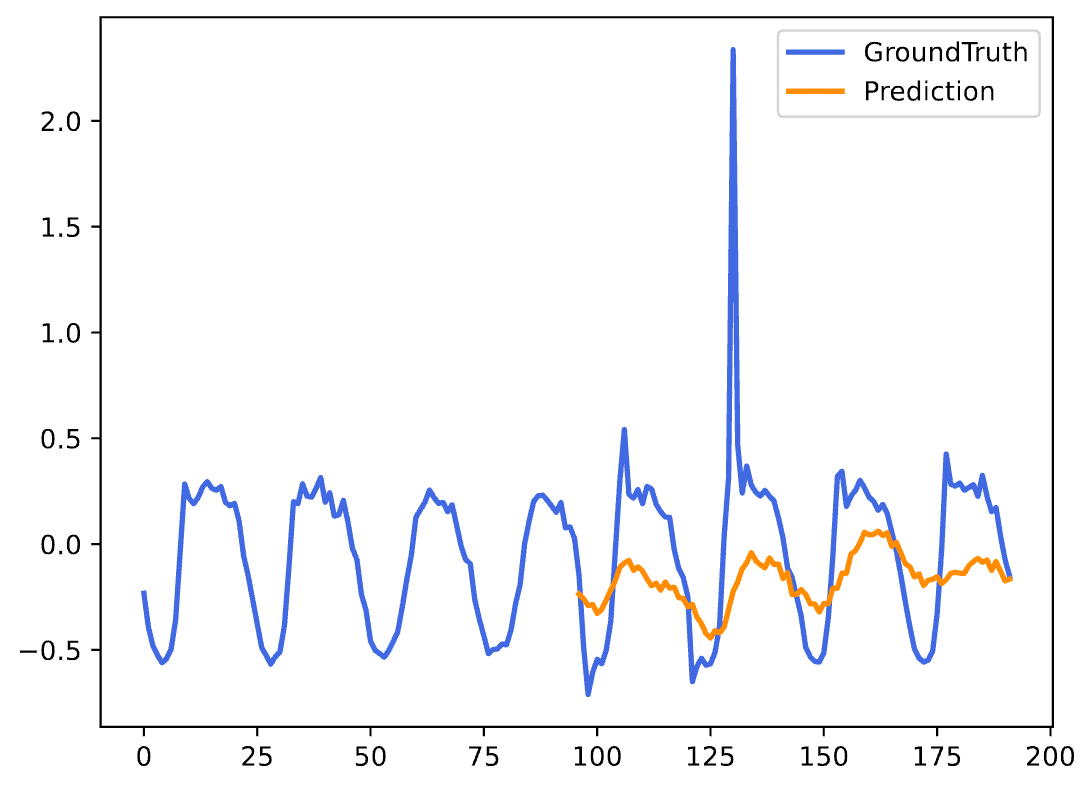}		
		\caption{CSDI}
	\end{subfigure}
	\caption{Case study on Traffic dataset. }
	\label{figure_case}
\end{figure}

We present actual prediction examples of three methods on the traffic dataset in Figure \ref{figure_case}. From the graph, it can be seen that our UniDiff performs better than the other two methods in predicting mutations. We believe this is due to the support of multimodality, which enables the model to be aware of unexpected situations. Although MCD-TSF also has multimodal capability, its inherent modal matching order limits its learning ability, resulting in poor robustness.

\section{Conclusion}

In this paper, we introduce UniDiff, a unified diffusion framework that addresses the challenge of multimodal time series forecasting by incorporating a novel parallel fusion module for the adaptive integration of data and a decoupled classifier-free guidance mechanism for fine-grained control over textual and temporal conditioning. Extensive experiments demonstrate that UniDiff achieves state-of-the-art performance across eight real-world datasets, consistently outperforming a wide range of strong baselines. Comprehensive ablation studies and further analyses confirm that our proposed components are crucial for this success, providing superior fusion capabilities and a practical balance between accuracy and computational efficiency. While future work could extend this framework to additional modalities, UniDiff provides a robust and flexible solution that sets a new performance benchmark for multimodal TSF, paving the way for more sophisticated and context-aware predictive models.



\bibliographystyle{IEEEtran}
\bibliography{TKDE_main}

\end{document}